\documentclass[letterpaper, 10 pt, conference]{ieeeconf}  

\IEEEoverridecommandlockouts                              

\usepackage{graphicx} 
\usepackage{import} 
\usepackage{caption} 
\usepackage{subcaption} 
\usepackage{amsmath} 
\usepackage{amssymb}  
\usepackage[
pdftitle={One Object at a Time: Accurate and Robust Structure From Motion for Robots},
pdfauthor={Aravind Battaje, Oliver Brock},
pdfkeywords={Perception-Action Coupling;Fixation;Eye movement;Structure From Motion;Robotics},
pdfsubject={Accepted at 2022 IEEE/RSJ International Conference on Intelligent Robots and Systems}]
{hyperref} 
\usepackage{color}	

\usepackage{fancyhdr}

\mathchardef\mhyphen="2D 

\title{\LARGE \bf
One Object at a Time:\\Accurate and Robust Structure From Motion for Robots
}

\author{Aravind Battaje$^{1, 2}$ and Oliver Brock$^{1, 2}$
\thanks{$^{1}$ Robotics and Biology Laboratory, Technische Universität Berlin} 
\thanks{$^{2}$ Science of Intelligence, Research Cluster of Excellence, Berlin}%
\thanks{We gratefully acknowledge funded provided by the Deutsche Forschungsgemeinschaft (DFG, German Research Foundation) under Germany’s Excellence Strategy – EXC 2002/1 “Science of Intelligence” – project number 390523135.}
}

\fancypagestyle{IEEECopyright}{
	\lfoot{\copyright2022 IEEE. Accepted at 2022 IEEE/RSJ International Conference on Intelligent Robots and Systems (IROS). \quad\quad\quad DOI: 10.1109/IROS47612.2022.9981953} 
}

\begin{document}
\maketitle
\thispagestyle{empty}
\pagestyle{empty}

\begin{abstract}


A gaze-fixating robot perceives distance to the fixated object and relative positions of surrounding objects immediately, accurately, and robustly. We show how fixation, which is the act of looking at one object while moving, exploits regularities in the geometry of 3D space to obtain this information. These regularities introduce rotation-translation couplings that are not commonly used in structure from motion. To validate, we use a Franka Emika Robot with an RGB camera. We a) find that error in distance estimate is less than 5 mm at a distance of 15 cm, and b) show how relative position can be used to find obstacles under challenging scenarios. We combine accurate distance estimates and obstacle information into a reactive robot behavior that is able to pick up objects of unknown size, while impeded by unforeseen obstacles. Project page: \url{https://oxidification.com/p/one-object-at-a-time/}.


\end{abstract}


\thispagestyle{IEEECopyright}

\section{Introduction}

\begin{figure}[t]
	\centering
	\includegraphics[width=0.75\columnwidth]{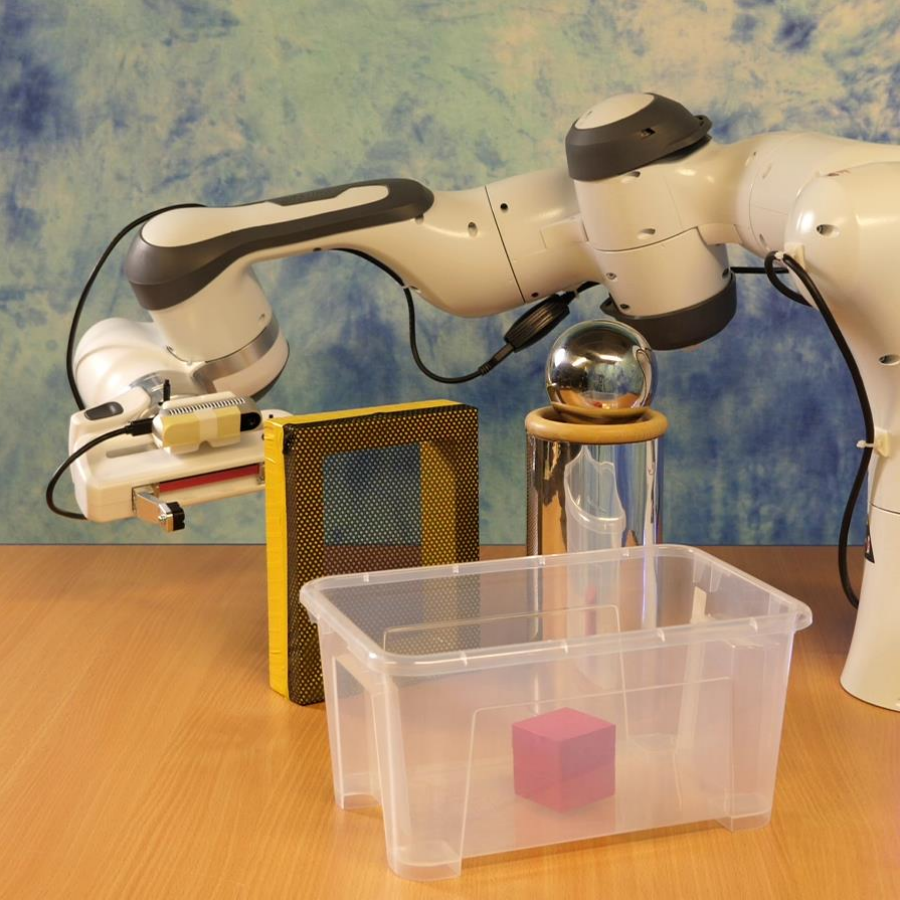}
	\caption{Problems for vision in robotics: reflective and translucent objects; we present an approach that robustly avoids these objects while picking up an object of unknown size, using only an RGB camera}
	\label{fig:intro_figure}
\end{figure}

When humans move about, they fixate their gaze on objects of interest without conscious notice. Some theories suggest it may be to stabilize images on the retina, or expose those objects to the area of retina with highest visual acuity (fovea). We show that this action of fixating on one object at a time simplifies perception for any moving agent. It gives \emph{robust} and \emph{instantaneous} information crucial to interact with the fixated object: how far is it and how other objects are positioned with respect to it. This information can be easily adapted to various robotic behaviors, including picking objects in challenging scenarios such as in Fig.~\ref{fig:intro_figure}.

The concept and method we describe in the paper apply to actively moving agents. So, we use the term `fixation' to mean fixating upon a point on an object \emph{while} moving.

Fixation gives a typical robot access to relevant information about interacting with the world. It provides a way to measure \textbf{accurate distance} to an object by only observing robot's self-movement, and robustly extract \textbf{relative positions} of surrounding objects by observing direction of visual motion on those objects. 

These properties are also efficient to extract. A robot measures its own movement to control its position which directly helps calculate distances, and simple image tracking can sense direction of visual motion, which helps find relative positions of other objects. Moreover, a robot only needs to modify the way it moves: fixating on one object at a time, instead of moving in an unconstrained manner. Consequently, this movement permits it to act robustly in the world without needing a world model.

In this paper, we show how distance estimation using fixation outperforms that of a translating camera, and how determining obstacles is immune to common problems for vision such as translucent and reflective surfaces. We integrate these accurate and robust properties enabled by fixation into a real-world robotic application. We task a robot to pick up an object of unknown size impeded by unforeseen and dynamic obstacles. The robot can be seen in action\footnote{A video is available at \url{https://youtu.be/o6xqkQINhmc}.\label{footnote:videos}} in Fig.~\ref{fig:robot_demo}.

\section{Related Work}

In our method, we actively control a moving robot's view in order to maintain fixation on one object at a time. By doing this, we extract relevant 3D properties of the world. So, our work relates to active vision, structure from motion (SfM) and other approaches that use camera fixations. In the following, we relate to these topics.

\subsection{Structure From Motion}
Our work lies in the intersection of SfM and active vision~\cite{aloimonos_active_1988}. Recovering 3D properties from a moving camera is well-posed and stable~\cite{aloimonos_active_1988} within active vision. But due to issues we explain below, depth recovery might not be accurate. Fixations can be seen as introducing constraints to overcome those issues.

Typical approaches in SfM recover depth from the component of visual motion due to camera translations, as rotations alone do not reveal depth cues. Simpson says, ``any process or mechanism for getting depth from optic flow must somehow filter out the rotational component''~\cite{simpson_optic_1993}. So, camera rotations are either removed~\cite{bideau_motion_2020} or compensated~\cite{zingg_mav_2010}. We show that camera rotations produced by fixation are indeed useful, especially in the context of robotics.

It is common in SfM to rely on point-features or ``corners''~\cite{hartley_multiple_2004}, to circumvent correspondence issues due to the aperture problem~\cite{simpson_optic_1993} at edge features. But corners are also problematic: there might not be enough corner features on smooth objects, or too many on textured objects that might overwhelm corner matching. In contrast to using corners, we simply rely on points at edges that are easier to extract, and obtain closest point correspondences on them. This is sufficient to determine location of other objects through their direction of visual motion.

\subsection{Fixation-Based Vision}
Some works~\cite{antonelli_bayesian_2014},~\cite{duran_robot_2020},~\cite{santini_active_2007} have demonstrated superior depth recovery by reproducing characteristic fixation movement of humans on robots. Similar to our approach, they take advantage of the robustness achieved with fixation. In contrast, instead of building a depth map for the full field-of-view, we only retrieve information that is necessary for the robot to immediately take action, i.e., knowledge of impeding objects and precise distance to the fixated object. We also show how these properties can be easily tailored to robust robot behaviors.

Fixation establishes a relationship between the scene and camera called the \emph{Zero Flow Circle (ZFC)}~\cite{raviv_quantitative_1991}, which can determine direction of heading~\cite{jain_peripheral_1996}. We too use the ZFC but to filter objects ``in front'' of the fixated object from the rest of the visual scene.


\section{Motion Parallax With Gaze Fixation}
\label{sec:motion_parallax_with_gaze_fixation}

In this section, we will look at the mechanics of gaze fixation, and how a camera moving in such a manner leads to robust properties.

\begin{figure}[t]
	\centering
	\begin{subfigure}{\columnwidth}
		\centering
		\def\svgwidth{\textwidth}
\begingroup%
  \makeatletter%
  \providecommand\color[2][]{%
    \errmessage{(Inkscape) Color is used for the text in Inkscape, but the package 'color.sty' is not loaded}%
    \renewcommand\color[2][]{}%
  }%
  \providecommand\transparent[1]{%
    \errmessage{(Inkscape) Transparency is used (non-zero) for the text in Inkscape, but the package 'transparent.sty' is not loaded}%
    \renewcommand\transparent[1]{}%
  }%
  \providecommand\rotatebox[2]{#2}%
  \newcommand*\fsize{\dimexpr\f@size pt\relax}%
  \newcommand*\lineheight[1]{\fontsize{\fsize}{#1\fsize}\selectfont}%
  \ifx\svgwidth\undefined%
    \setlength{\unitlength}{645.99583423bp}%
    \ifx\svgscale\undefined%
      \relax%
    \else%
      \setlength{\unitlength}{\unitlength * \real{\svgscale}}%
    \fi%
  \else%
    \setlength{\unitlength}{\svgwidth}%
  \fi%
  \global\let\svgwidth\undefined%
  \global\let\svgscale\undefined%
  \makeatother%
  \begin{picture}(1,0.53095099)%
    \lineheight{1}%
    \setlength\tabcolsep{0pt}%
    \put(0,0){\includegraphics[width=\unitlength,page=1]{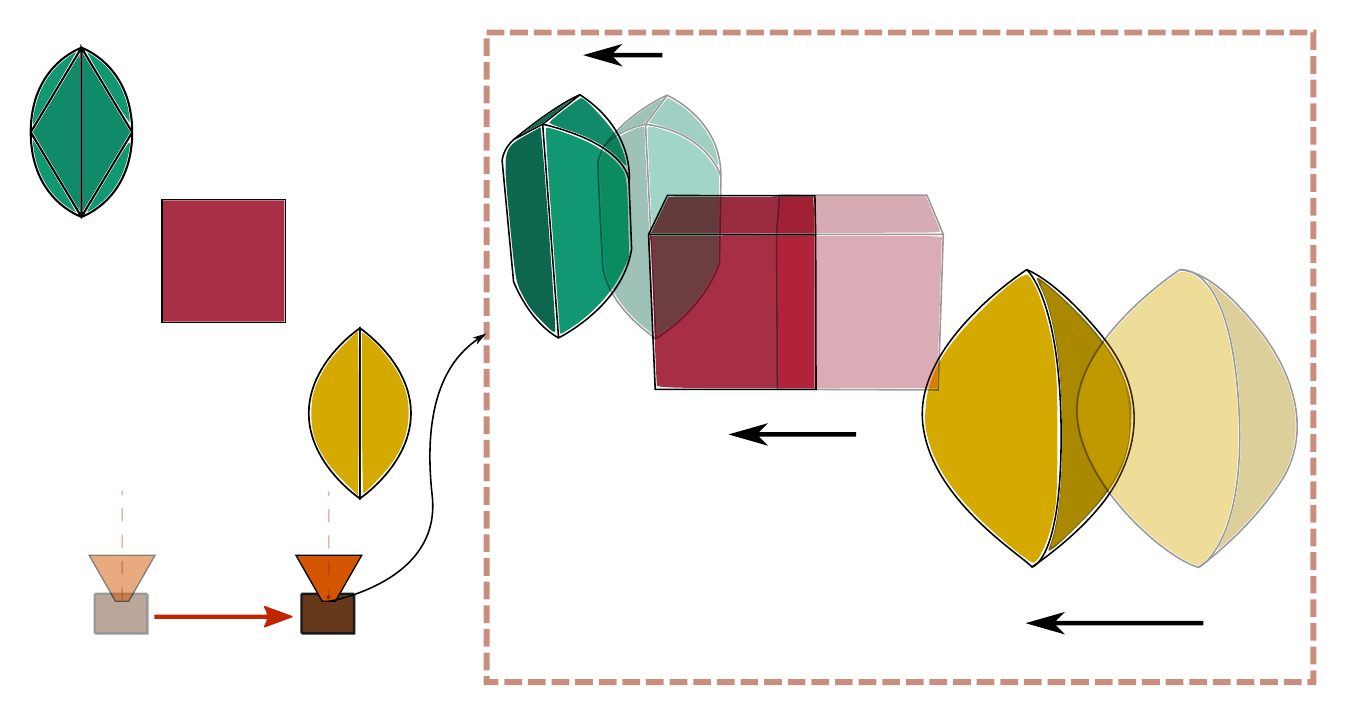}}%
    \put(0.13070347,0.47435017){\color[rgb]{0,0,0}\makebox(0,0)[lt]{\lineheight{1.25}\smash{\begin{tabular}[t]{l}Top-view\end{tabular}}}}%
    \put(0.7448455,0.47435017){\color[rgb]{0,0,0}\makebox(0,0)[lt]{\lineheight{1.25}\smash{\begin{tabular}[t]{l}Camera-view\end{tabular}}}}%
    \put(0.1303709,0.43924597){\color[rgb]{0,0,0}\makebox(0,0)[lt]{\lineheight{1.25}\smash{\begin{tabular}[t]{l}(not to scale)\end{tabular}}}}%
  \end{picture}%
\endgroup%

		\caption{Motion parallax with a translating camera}
		\label{fig:top_and_camera_view_nogazelock}
	\end{subfigure}	
	\begin{subfigure}{\columnwidth}
		\centering
		\def\svgwidth{\textwidth}
\begingroup%
  \makeatletter%
  \providecommand\color[2][]{%
    \errmessage{(Inkscape) Color is used for the text in Inkscape, but the package 'color.sty' is not loaded}%
    \renewcommand\color[2][]{}%
  }%
  \providecommand\transparent[1]{%
    \errmessage{(Inkscape) Transparency is used (non-zero) for the text in Inkscape, but the package 'transparent.sty' is not loaded}%
    \renewcommand\transparent[1]{}%
  }%
  \providecommand\rotatebox[2]{#2}%
  \newcommand*\fsize{\dimexpr\f@size pt\relax}%
  \newcommand*\lineheight[1]{\fontsize{\fsize}{#1\fsize}\selectfont}%
  \ifx\svgwidth\undefined%
    \setlength{\unitlength}{643.43160794bp}%
    \ifx\svgscale\undefined%
      \relax%
    \else%
      \setlength{\unitlength}{\unitlength * \real{\svgscale}}%
    \fi%
  \else%
    \setlength{\unitlength}{\svgwidth}%
  \fi%
  \global\let\svgwidth\undefined%
  \global\let\svgscale\undefined%
  \makeatother%
  \begin{picture}(1,0.53306695)%
    \lineheight{1}%
    \setlength\tabcolsep{0pt}%
    \put(0,0){\includegraphics[width=\unitlength,page=1]{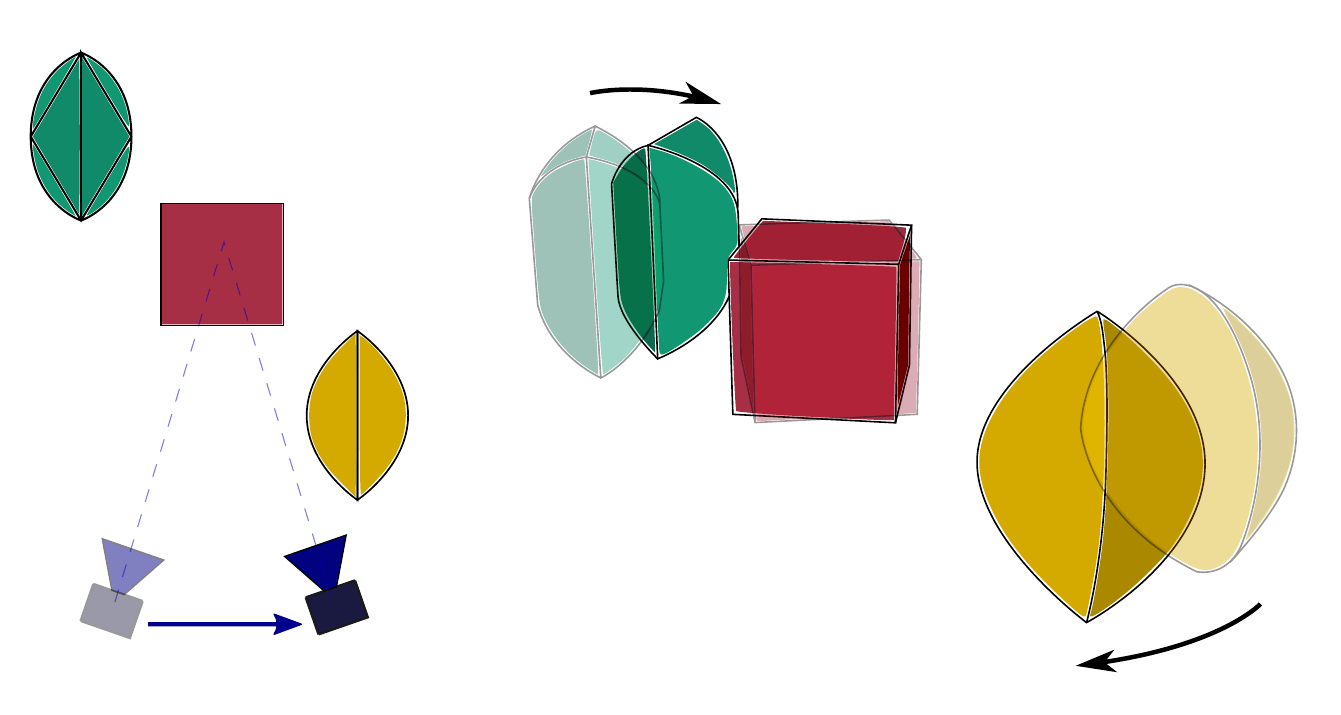}}%
    \put(0.13122436,0.47624057){\color[rgb]{0,0,0}\makebox(0,0)[lt]{\lineheight{1.25}\smash{\begin{tabular}[t]{l}Top-view\end{tabular}}}}%
    \put(0.74781388,0.47624057){\color[rgb]{0,0,0}\makebox(0,0)[lt]{\lineheight{1.25}\smash{\begin{tabular}[t]{l}Camera-view\end{tabular}}}}%
    \put(0.13089045,0.44099647){\color[rgb]{0,0,0}\makebox(0,0)[lt]{\lineheight{1.25}\smash{\begin{tabular}[t]{l}(not to scale)\end{tabular}}}}%
    \put(0,0){\includegraphics[width=\unitlength,page=2]{combined_gazelock.pdf}}%
  \end{picture}%
\endgroup%

		\caption{Motion parallax with a moving camera fixating on the (center) red object. Due to fixation, the camera translates and rotates}
		\label{fig:top_and_camera_view_gazelock}
	\end{subfigure}	
	\caption{Translating camera leads to parallax only in one direction, whereas gaze fixation leads to parallax in different directions. Translucent colors indicate camera in starting position, and solid colors in the final position. \emph{(a)}~Farther objects move slower. Thus distance to an object can be found by measuring object translation for a given amount of camera translation. \emph{(b)}~Fixation produces motion in objects directly indicative of their position: fixated object moves minimally, objects behind this move in the opposite direction as objects in front. Distance to fixated object is the ratio of camera rotation to translation.}
	\label{fig:gaze_and_nogazelock_schematic}
\end{figure}

In order to understand fixation, let us first look at the effects of camera translation and rotation independently. Camera translation results in motion parallax as shown in Fig.~\ref{fig:top_and_camera_view_nogazelock}. Here, parallax is the apparent change in parts of the scene on the camera sensor surface, and distant objects produce lesser parallax. In contrast, camera rotations produce the same parallax for any object irrespective of distance.

Fixation combines camera translation and rotation in a unique way. In order to maintain an object in the same position in camera's view, fixation produces a rotation that counteracts the effect of translation. In other words, the parallax due to rotation cancels that of translation. As a consequence, objects behind the fixated object produce parallax in the opposite direction as those in front, as seen in Fig.~\ref{fig:top_and_camera_view_gazelock}.

This unique combination of translation and rotation produces robust information that is immediately helpful. The parallax in opposing directions serves as a filter to quickly \textbf{distinguish between objects that are in front, and behind the fixated object}. The rotations are also dependent on distance to the object: farther objects induce lesser rotations to maintain fixation. Thus, the ratio between rotation and translation immediately reveals the \textbf{distance to the fixated object}.

We use these two principles as the basis to generate behaviors in a robot in Sec.~\ref{sec:fixation_for_robots}. In this section, we explore these principles in more detail. The following two subsections explain the mathematical underpinnings, by first taking a look at the general case of camera motion, and later analyzing the influence of fixation.

\subsection{Visual Motion Patterns Due to Camera Motion} 
Let us understand the general mathematical relation that expresses motion parallax. A camera projects 3D points onto the 2D sensor surface, and for a moving camera, parallax can be expressed as changes of those projected points for changes of camera translation and rotation \cite{simpson_optic_1993}, \cite{chaumette_visual_2006}.

The standard pinhole projection model relates a 3D point in the camera frame with coordinates $[X, Y, Z]^T$ to its 2D projection on the image plane $[x, y]^T$. This relation is given by the projection equations $x = f \cdot X/Z$ and $y = f \cdot Y/Z$. In rest of the paper, the focal length $f$ is assumed to be unit length without loss of generality.

For the case of a moving camera, the velocity of projected points $[\dot{x}, \dot{y}]^T$ can be related to the camera velocity with the help of time derivative of projection equations. Eqn. \ref{eqn:basic_eqn} shows this relationship. The reader can find more information about this derivation in \cite{chaumette_visual_2006}.

\begin{equation}
	\begin{split}
		\label{eqn:basic_eqn}
		\begin{bmatrix}
			\dot{x} \\
			\dot{y}
		\end{bmatrix}
		=
		\phantom{lotsofspace}
		d \cdot
		\begin{bmatrix}
			-1 & 0 & x \\
			0 & -1 & y
		\end{bmatrix}
		&\begin{bmatrix}
			v_X \\
			v_Y \\
			v_Z
		\end{bmatrix}
		+
		\\ 
		\begin{bmatrix}
			xy & -(1 + x^2) & y \\
			(1 + y^2) & -xy & -x
		\end{bmatrix}
		&\begin{bmatrix}
			\omega_X \\
			\omega_Y \\
			\omega_Z
		\end{bmatrix}
	\end{split}
\end{equation}
where $[v_X, v_Y, v_Z]^T$ is the camera translation velocity, $[\omega_X, \omega_Y, \omega_Z]^T$ is the camera rotation velocity, and $d = 1/Z$ is the disparity---the only quantity that relates to distance to 3D points. Since $d$ influences the first term, often camera translation is regarded as helpful in extracting 3D properties, and rotations be filtered out~\cite{simpson_optic_1993}. On the contrary, we show that camera rotations are indeed useful.

\subsection{Visual Motion Patterns Due to Gaze Fixation}
\label{subsec:influence_of_fixation}
Fixation can be seen as coupling two rotational degrees-of-freedom (DOF) with translations. So specifying camera velocities reduces to 4-DOF, instead of 6-DOF. This leads to a useful reduction of Eqn.~\ref{eqn:basic_eqn} that results in patterns of parallax explained earlier. And the coupling relates to the distance of fixated object. 

This coupling imposed by fixation is expressed as
\begin{align}
	\omega_X &= -v_Y \cdot d_f
	\label{eqn:substitue_omega_X}\\
	\omega_Y &= \phantom{+}v_X \cdot d_f 
	\label{eqn:substitue_omega_Y}
\end{align}
where $d_f = 1/Z_f$. This coupling entails camera rotation for any lateral translation to ensure a particular 3D point $[0, 0, Z_f]^T$ in the camera frame is held at the center of FOV $[0, 0]^T$. Please note, all quantities in this paper with subscript $f$ denote their association with the \emph{fixation point}. Therefore, $Z_f$ is the distance to the fixated point.

For clear exposition, let us also set $v_Z = 0$ and $\omega_Z = 0$. The former leads to a quantity called time-to-collision (TTC)~\cite{simpson_optic_1993}, which is not related to motion parallax. And the latter does not couple with fixation. In practice both can be accounted. Thus, setting irrelevant quantities to zero, and substituting Eqn. \ref{eqn:substitue_omega_X} and \ref{eqn:substitue_omega_Y} in Eqn. \ref{eqn:basic_eqn} produces
\begin{equation}
	\label{eqn:gaze_fixed_eqn}
	\begin{bmatrix}
		\dot{x} \\
		\dot{y}
	\end{bmatrix}
	= \\
	\begin{bmatrix}
		(1 + x^2) d_f - d & -x y \, d_f \\
		xy \, d_f & (1 + y^2) d_f - d
	\end{bmatrix}
	\begin{bmatrix}
		v_X \\
		v_Y
	\end{bmatrix}
\end{equation}

To understand the geometry induced by this expression, let us analyze the surface where projected velocities is zero, ie., $[\dot{x}, \dot{y}]^T = 0$. This is a surface that contains the point of fixation, but it also separates the space such that 3D points beyond this surface have the opposite sign of motion as those within the surface. This is also called the \emph{Zero Flow Circle}~\cite{raviv_quantitative_1991}.

For a robot interacting with the world, this separation is useful: based only on direction of motion, it can instantaneously determine parts of the scene ``behind'' the fixated point that can be simply ignored as far as direct interaction is concerned. Likewise, it can instantaneously determine parts of the scene that is ``in front'' of the fixated point that could perhaps be an obstacle.

In practice, we can make an approximation to Eqn. \ref{eqn:gaze_fixed_eqn}. If the camera has a small FOV, then the quadratic terms quickly vanish to zero: $x^2, y^2, xy \approx 0$. Consequently, 
\begin{equation}
	\label{eqn:approx_gaze_fixed_eqn}
	\begin{bmatrix}
		\dot{x} \\
		\dot{y}
	\end{bmatrix}
	\approx \\
	\begin{bmatrix}
		d_f - d & 0 \\
		0 & d_f - d
	\end{bmatrix}
	\begin{bmatrix}
		v_X \\
		v_Y
	\end{bmatrix}
\end{equation}
We use this approximation to construct a potential field for obstacles in Sec. \ref{subsec:avoiding_obstacles} by simply filtering points that have the opposite sign of camera movement (effectively $d_f < d$).

In the analysis so far, we introduced a new coupling (Eqn. \ref{eqn:substitue_omega_X}, and \ref{eqn:substitue_omega_Y}), but it need not be pre-computed. Instead, the coupling comes about as a \emph{result} of fixation. Fixation forces the camera to follow circles of different radius, depending on the distance to that 3D point of interest. The robot can then simply determine the instantaneous radius of curvature by examining its ratio of translation and angular velocities.

In summary, Eqn.~\ref{eqn:gaze_fixed_eqn} (or Eqn.~\ref{eqn:approx_gaze_fixed_eqn})) describes the phenomenon of parallax of opposite directions we saw in the beginning of this section, and the coupling described in Eqn. \ref{eqn:substitue_omega_X} and \ref{eqn:substitue_omega_Y} helps measure distance to fixated object.


\section{Gaze Fixation for Robust Robot Behavior}
\label{sec:fixation_for_robots}

Now let us look at the requirements for a robot to leverage the 3D properties enabled by fixation, and subsequently pick up objects of unknown size while avoiding obstacles. Later in this section, we explain our method that satisfies these requirements.

For purposes of simplicity, we will assume a single robot manipulator with a monocular camera sensor near its end-effector. The requirements we list below can be extended when camera can be moved independent of the manipulator.

The robot must 
\begin{enumerate}
	\item accomplish \emph{fixation} by actively rotating in order to keep a target object in the center of FOV; \label{reqitem:gaze-fix}
	\item translate \emph{laterally} to produce the special geometry induced by fixation; \label{reqitem:lateral-trans}
	\item \emph{estimate distance} to target by measuring the ratio of translation and rotation; \label{reqitem:meas-move}
	\item observe \emph{direction of movement} of objects relative to the target object; \label{reqitem:track-motion}
	\item \emph{avoid obstacles} informed by the relative visual motion;\label{reqitem:obst-trans}
	\item translate \emph{towards} the target object to pick it up;\label{reqitem:approach}
	\item and, optionally, place the end-effector in a desirable configuration to facilitate successful grasp. \label{reqitem:ee-config}
\end{enumerate}

To turn these requirements into implementation, we use an instantiation of visual servoing~\cite{chaumette_visual_2006} for ~\ref{reqitem:gaze-fix}, \ref{reqitem:lateral-trans}, \ref{reqitem:obst-trans}, and~\ref{reqitem:approach}. This is not a generic visual servoing, but one that structures perceptual processes enabled by fixation~\ref{reqitem:gaze-fix}. Then to enable reliable distance measurements~\ref{reqitem:meas-move}, we integrate the robot state using an Extended Kalman Filter. And for~\ref{reqitem:track-motion}, we use a simple form of optic flow. In the following subsections, we will explain our approach to visual servoing, optic flow, distance estimation and its use to avoid obstacles.

\subsection{Behavior Generation Based on Gaze Control}
\label{subsec:gcvs}

We partition the 6-DOF control of a robot into three components as detailed below.

\subsubsection{Gaze fixation}

This maintains the location of a specified feature $[x_s, y_s]^T$ in the center of the image using the following servo law:

\begin{equation}
	\label{eqn:vs_gazelock}
	\boldsymbol{\omega}_{XY} = \lambda_{XY} 
	\begin{bmatrix}
		xy & -(1 + x^2) \\
		(1 + y^2) & -xy
	\end{bmatrix}^{-1}
	\boldsymbol{e}_{xy}
\end{equation} 
where $\boldsymbol{\omega}_{XY} = [\omega_X, \omega_Y]^T$, and $\lambda_{XY}$ is the proportional gain to minimize error $\boldsymbol{e}_{xy} = [x_s, y_s]^T$. The feature $[x_s, y_s]^T$ may be any representative point within the area constituting an object. We use the centroid of the area occupied by a color-specified object as this feature\footnote{This simple scheme also provides a stable fixation point for hollow objects.}. Finally, the Jacobian matrix is always invertible, which ensures stable fixation and robust fulfillment of requirement~\ref{reqitem:gaze-fix}.

\subsubsection{Approach}
\label{subsec:approach_control}

The approach servo law brings the robot to some specified distance in front of the fixated object:
\begin{equation}
	\label{eqn:vs_approach}
	v_Z = \lambda_Z log \left( \frac{Z_d}{Z_f} \right)
\end{equation}
where $\lambda_Z$ specifies how fast to approach or retract from desired distance $Z_d$, and $Z_f$ is the estimated fixation distance (estimation is explained in Sec. \ref{subsec:online_distance_estimation}).

This servo law satisfies requirement \ref{reqitem:approach} to move toward the object. To pick the target object up, we employ an open-loop maneuver: move the end-effector forward by $Z_d$ and close the gripper.

\subsubsection{Translate laterally}
\label{subsec:lateral_translation}
Lateral translations of the robot once an object is fixated serves two objectives: produce visual motion (requirement~\ref{reqitem:lateral-trans}) and depending on existence of obstacles, move away from them (requirement~\ref{reqitem:obst-trans}). The following control law accomplishes both:
\begin{equation}
	\label{eqn:vs_our_heuristic}
	\boldsymbol{v}_{XY} = \boldsymbol{\lambda}_{dist} (Z_f - Z_d) + 
	\begin{bmatrix}
		\lambda_{cycle_X} sin(t) \\
		\lambda_{cycle_Y} cos(t)
	\end{bmatrix} -
	\lambda_{obs}\boldsymbol{obs}_{xy}
\end{equation}
where $\boldsymbol{v}_{XY} = [v_X, v_Y]^T$, and the three terms in control law are independent heuristics, weighted by their respective $\lambda$ gains. The first term moves the robot laterally if it is far from the target, or if estimated distance to target is uncertain. The second term introduces small periodic movements. These two terms ensure motion needed to constantly determine distance to target object, and relative location of other objects. The last term moves the robot away from obstacles (more details in Sec. \ref{subsec:avoiding_obstacles}).

\hfill \break
The above three components are reactive control laws that hold the target object as temporary reference even when it moves, or obstacles change positions. Thus they result in a rich and robust behavior\footnote{Note, we leave out control of $\omega_Z$. If desired, this maybe used in conjunction with an additional heuristic to Eqn. \ref{eqn:vs_our_heuristic}, to place the end-effector in some desirable configuration for pick up (requirement \ref{reqitem:ee-config}).}.


\subsection{Optic flow (requirement \ref{reqitem:track-motion})}
Optic flow algorithms enable tracking visual motion of points on objects. This helps determine position of objects relative to fixated object. However, since only direction of movement is important, the optic flow algorithm can be simple. We use optic flow derived from spatiotemporal gradients~\cite{simoncelli_distributed_1993},~\cite{simoncelli_design_1994}.

\subsection{Online Distance Estimation (requirement \ref{reqitem:meas-move})}
\label{subsec:online_distance_estimation}

We employ an Extended Kalman Filter (EKF) to maintain a temporally consistent estimate of fixation point and associate uncertainty to it. The EKF filters over camera velocity and inverse distance to fixation point $d_f$. Its motion model integrates velocity with constant acceleration, and implements the fixation constraint from Eqn. \ref{eqn:substitue_omega_X} and \ref{eqn:substitue_omega_Y}. The EKF's observation model integrates camera velocity (obtained from robot's velocity) along with $n$ samples of optic flow on the target object. We randomly sample $n = 30$ points on edge features of the color-specified object. The randomness allows sustained estimation with drastic changes in perspective.

The EKF also recovers quickly if the target moves, or the robot is unable to maintain fixation. As fixation maintains near-zero optic flow on the target object, any large deviations from it indicates inconsistencies, thus increasing uncertainty of estimate.


\subsection{Avoiding Obstacles (aiding requirement \ref{reqitem:obst-trans})}
\label{subsec:avoiding_obstacles}
As the presence of other objects ``in front'' of fixated object becomes apparent immediately through its direction of movement, this can instruct the robot to produce lateral translations (Sec. \ref{subsec:lateral_translation}) to move in a direction with least influence of those objects. To enable this, we construct a potential function populated with high values for any pixels that have optic flow in the opposite sign of robot lateral movement, i.e., $[\dot{x}, \dot{y}]\cdot[v_X, v_Y]^T < 0$. We then calculate the center-of-mass of this potential field as $\boldsymbol{obs}_{xy}$, and instruct the robot to move in the \emph{opposite} direction of it.


\section{Experiments}
\subsection{Experimental Setup}
Here we describe the platform and parameters we use for all the upcoming sections. We use a Franka Emika Robot with a RealSense D435 camera rigidly attached near the end-effector---we only use the RGB sensor on the D435. The robot is controlled using the method described in previous section, in realtime at 1000 Hz. The optic flow and distance estimation run at approximately 25 Hz. We run all the above on a compact PC with an Intel Core i5 CPU.

\subsection{Gaze Fixation Estimates Distances Accurately}
In this subsection, we evaluate and compare distance estimates from a fixating robot and a translating robot. This is similar to Fig. \ref{fig:top_and_camera_view_gazelock} and \ref{fig:top_and_camera_view_nogazelock} respectively.

%
%
%

\begin{figure}[h]
	\centering
	\begin{subfigure}{\columnwidth}
		\centering
		\includegraphics[width=0.75\columnwidth]{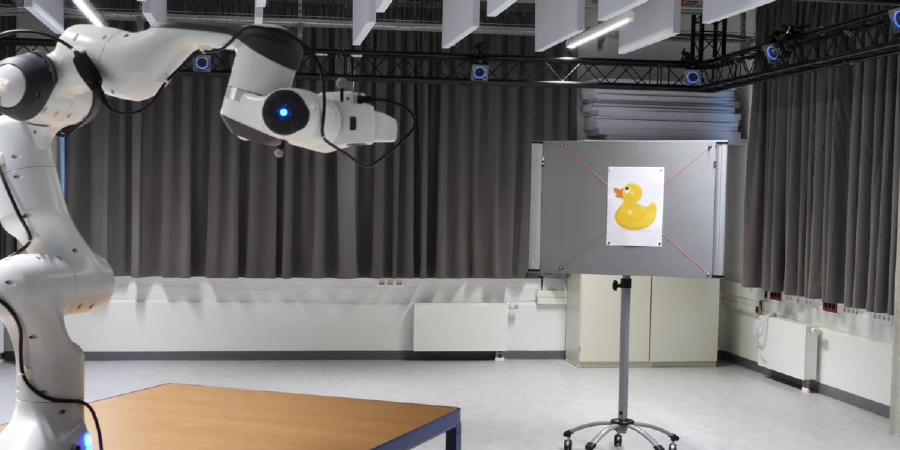}
		\caption{Robot directed at a target in a motion capture system}
		\label{fig:dist_eval_setup}
	\end{subfigure}
	
	\begin{subfigure}{\columnwidth}
		\includegraphics[width=\columnwidth]{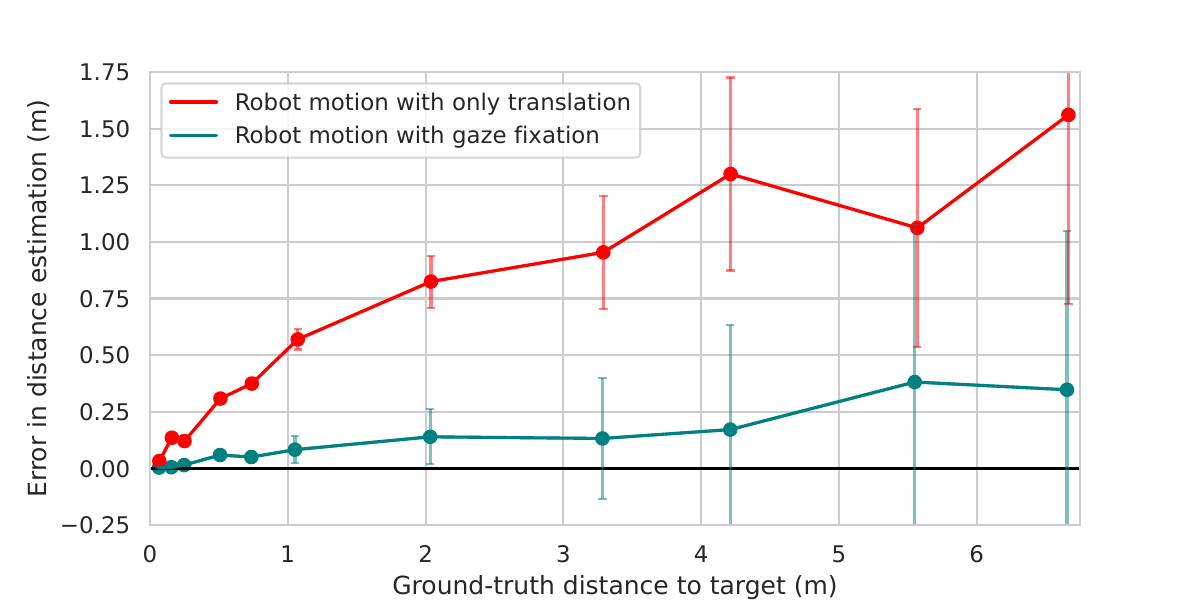}
		
		\caption{Distance estimation of a fixating robot outperforms that of a translating robot; Note, for visual clarity, we have flipped the sign of error for distance estimates under fixation}
		
		\label{fig:dist_eval}
	\end{subfigure}
	\caption{Distance estimation experimental setup and results}
\end{figure}

The robot estimates distance to a target without moving toward it. Hence we disable approach (Sec.~\ref{subsec:approach_control}). The target is a printed ``rubber duck''. These targets are printed at 5 different sizes to prevent the target from being too small ($<5\%$ of image width) or too large ($>75\%$ of image width) in the camera view at different distances. Our method has no information of the size of target. We vary the target distance two orders of magnitude, from 66 mm to 6.6 m. At each target distance, the robot end-effector translates laterally in a cyclic motion. Its amplitude is the same for both fixation and translation-only motion. To estimate distances, we use the EKF from Sec.~\ref{subsec:online_distance_estimation}. In the case of translation-only motion, we simply remove the \emph{fixation constraint} in its motion model. To obtain the ground truth distances, the positions of robot end-effector and target are tracked in a high-precision motion capture system from OptiTrack, as shown in Fig.~\ref{fig:dist_eval_setup}. At each target distance, we report the mean and standard deviation of error between ground truth and estimated data from EKF, accumulated over approximately 9 seconds (225 samples).

Fig. \ref{fig:dist_eval} shows results of the evaluation. It shows that distance estimation error with fixation is significantly less than that when robot only translates. The error is less than 5 mm at around 15 cm distance---this would be a reasonable distance for a manipulator to start interacting with an object, e.g., pick it up. The error also grows slower with larger distances: it is less than 13 cm at around 2 m, which is only about 15\% of translation-only motion. These impressive results indicate that accurate distances using fixation can be utilized in real-world robots using only an RGB camera.

\subsection{Gaze Fixation Is Robust in Challenging Scenarios}
\label{subsec:challening_scenarios}
\begin{figure*}
	\centering
	\includegraphics[width=1.0\textwidth]{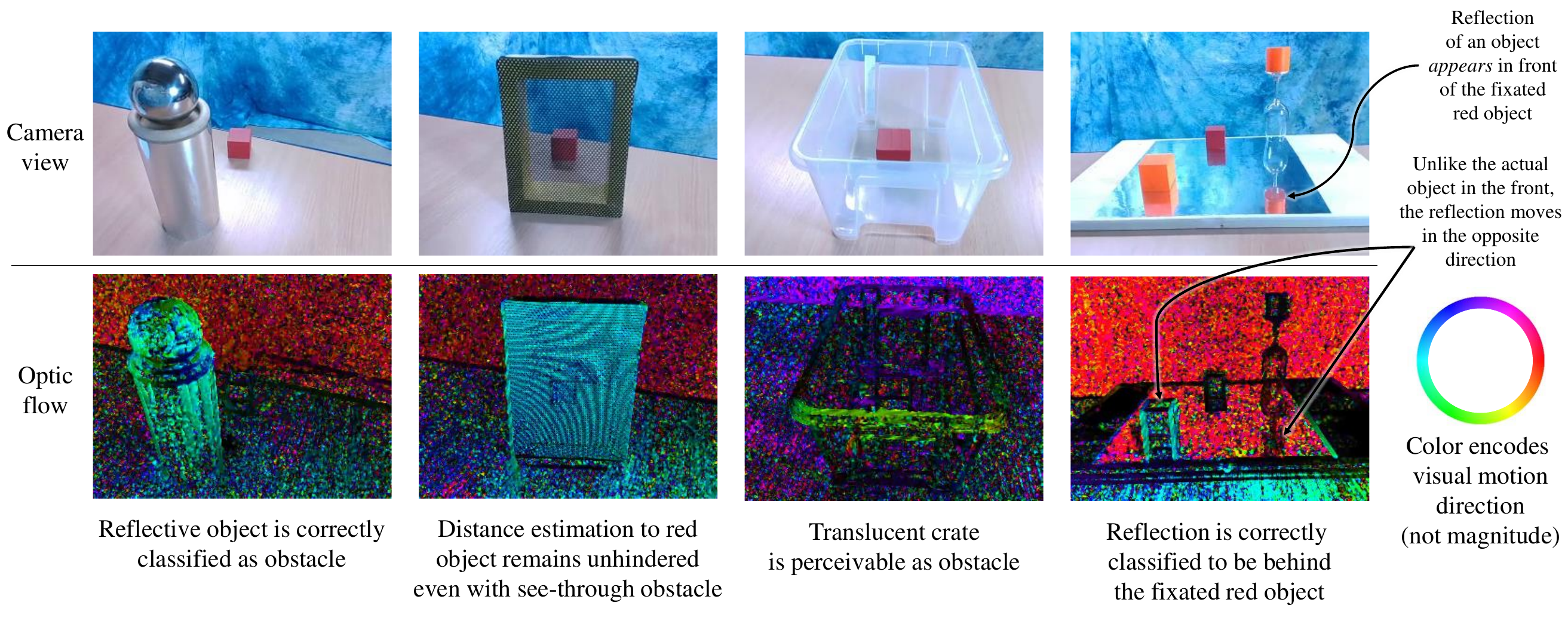}
	\caption{Four scenarios difficult for vision but easy with fixation (more details in Sec.~\ref{subsec:challening_scenarios})}
	\label{fig:rgb_and_opticflow_special_situations}
\end{figure*}
We will now examine how fixation leads to robust information about surrounding objects \emph{while} estimating accurate distances to the fixated object. Fig.~\ref{fig:rgb_and_opticflow_special_situations} shows the camera view and optic flow visualization for different scenarios. In each, the robot fixates on the red object. The opposing colors in optic flow visualization distinctly indicate relative positions of objects and scene surrounding the red fixated object---all objects in the front move in the opposite direction of those in the back. Even object reflections (fourth column) that appear to be in front can be clearly dismissed, because reflections move as if they were behind the fixated object. Finally, even though an obstacle might be physically in the line-of-sight of the fixated object (second column), distance to red object can be calculated independently as this only depends on the object itself being visible in camera view. As extracting this information requires only a basic optic flow implementation and no prior object models, this may be effortlessly integrated into robust robotic applications.

\subsection{Gaze Fixation Produces Robust Robot Behavior}
\begin{figure}[h]
	\centering	
	\includegraphics[width=0.75\columnwidth]{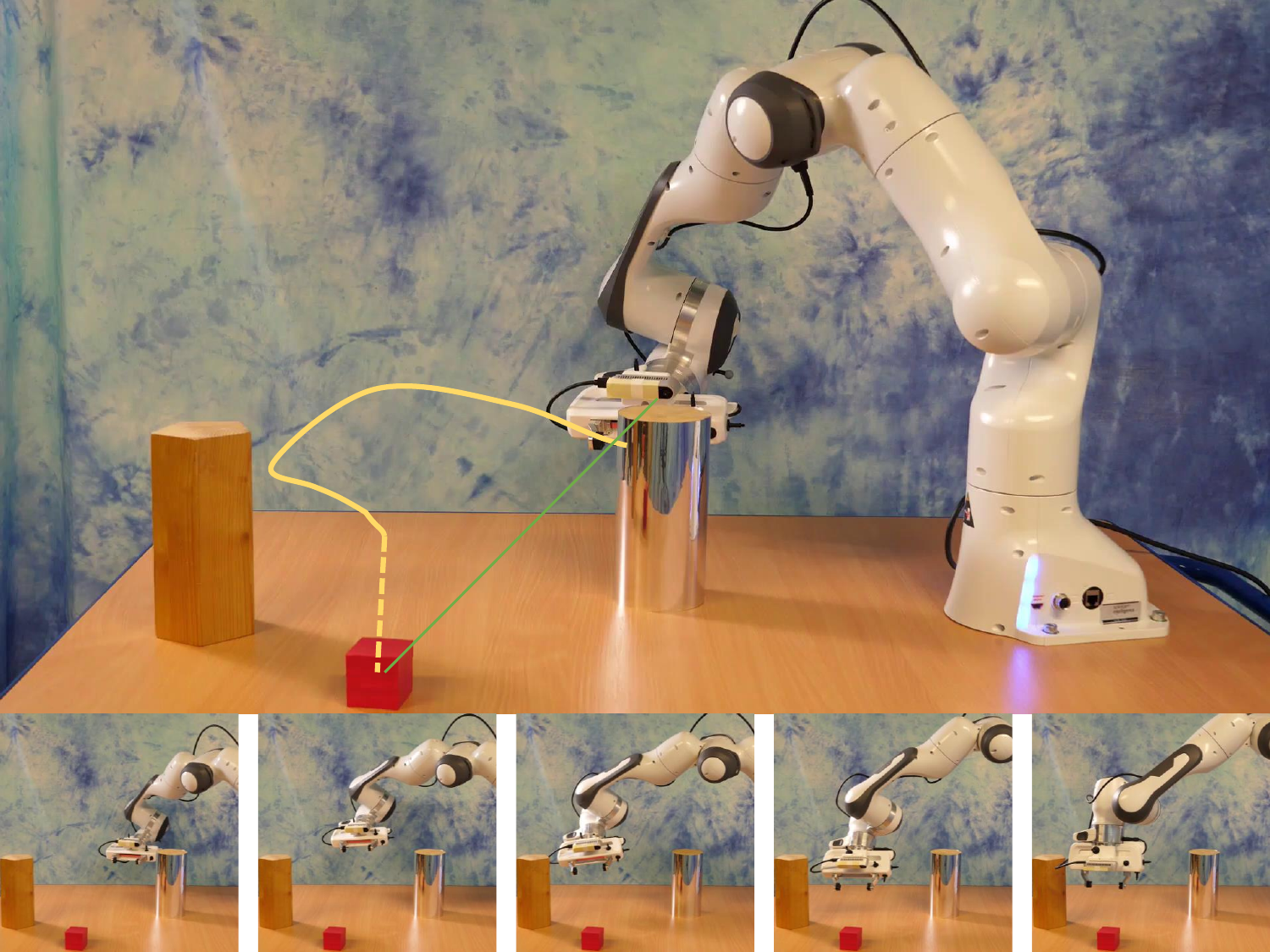}
	\caption{Robot takes a complex path (yellow) to fetch the red object, in spite of several obstacles. The robot begins to avoid the reflective obstacle in its line-of-sight (green), and then avoids the wooden obstacle in the far left upon discovering it. Finally, the robot moves forward (dotted yellow) in an open loop fashion to grasp the red object.}
	\label{fig:robot_demo}
\end{figure}

As we described in the previous subsections, using fixation, the robot can quickly extract robust information about surrounding objects, while estimating accurate distances to fixated object. We also described in Sec.~\ref{sec:fixation_for_robots} an approach to combine this information and generate robot behavior. We demonstrate this on a robot\textsuperscript{~\ref{footnote:videos}} as shown in Fig.~\ref{fig:robot_demo}. Here the robot succeeds in picking up the target object, even when impeded by obstacles from Fig.~\ref{fig:rgb_and_opticflow_special_situations}, or if obstacles are moved mid-operation, or the target itself is moved. This leads us to conclude that fixation is a powerful behavior to incorporate in a robot, especially to build other robust behaviors on top of it.


\section{Limitations}

\paragraph*{Dependence on rotation} With fixation, distance recovery is dependent on amount of camera rotation: farther objects result in lesser rotations. Hence fidelity in measuring small camera rotation is paramount for farther objects. Without this, distance recovery will be erroneous. However, determining relative location of other objects does not impose restrictions on measuring camera rotations.

\paragraph*{Stability of fixation}
Unstable fixation point, e.g., due to noisy object detection, produces spurious camera rotations. Consequently distance estimation might not converge.

\paragraph*{Ambiguity in fixation}
We classify an object of interest using its color. This results in ambiguous fixation if there are other objects with the same color. This can be resolved with sophisticated object detectors and visual tracking.

\paragraph*{Simplicity of obstacle avoidance}
We employ a simple heuristic to avoid obstacles based on a potential function. This provides a powerful local motion primitive, but is prone to local minima. To avoid this, a motion planner could be used in conjunction to express even richer behaviors. We have also ignored the contribution of TTC to this potential function, thus we maintain the approach of the robot relatively slower than lateral translations. TTC can be accounted as it results in a shifted ZFC~\cite{raviv_quantitative_1991} that still contains the fixation point.

\section{Conclusions}
Gaze fixation leverages useful relationships between a moving agent and its surrounding world: it gives just-in-time access to distance to a fixated object, and relative location of other objects. We show how these properties emerge from fixation, and show that the resultant distance estimates are accurate and information about surrounding objects are robust. We also demonstrate a way to generate robot behavior utilizing these properties.

Fixation is also easy and efficient. It adds rotations that are not adversarial to a typical robot's objective; uses robot's proprioception to measure distance; and utilize simple visual tracking to find relevant details of surrounding objects. It is also easy to adapt these properties into robust behaviors. We believe the perceptual and behavioral skills offered in this paper form a foundation for generating robust real-world robotic applications. We believe fixation has the potential to become ubiquitous in robotics.


\bibliographystyle{IEEEtran}
\bibliography{Battaje-22-IROS}

\begin{thebibliography}{10}
\providecommand{\url}[1]{#1}
\csname url@rmstyle\endcsname
\providecommand{\newblock}{\relax}
\providecommand{\bibinfo}[2]{#2}
\providecommand\BIBentrySTDinterwordspacing{\spaceskip=0pt\relax}
\providecommand\BIBentryALTinterwordstretchfactor{4}
\providecommand\BIBentryALTinterwordspacing{\spaceskip=\fontdimen2\font plus
\BIBentryALTinterwordstretchfactor\fontdimen3\font minus
  \fontdimen4\font\relax}
\providecommand\BIBforeignlanguage[2]{{%
\expandafter\ifx\csname l@#1\endcsname\relax
\typeout{** WARNING: IEEEtran.bst: No hyphenation pattern has been}%
\typeout{** loaded for the language `#1'. Using the pattern for}%
\typeout{** the default language instead.}%
\else
\language=\csname l@#1\endcsname
\fi
#2}}

\bibitem{aloimonos_active_1988}
J.~Aloimonos, I.~Weiss, and A.~Bandyopadhyay, ``\BIBforeignlanguage{en}{Active
  vision},'' \emph{\BIBforeignlanguage{en}{Intl. Journal of Computer Vision}},
  vol.~1, no.~4, pp. 333--356, Jan. 1988.

\bibitem{simpson_optic_1993}
W.~A. Simpson, ``\BIBforeignlanguage{en}{Optic flow and depth perception},''
  \emph{\BIBforeignlanguage{en}{Spatial Vision}}, vol.~7, no.~1, pp. 35--75,
  1993.

\bibitem{bideau_motion_2020}
P.~K. Bideau, ``Motion {Segmentation} - {Segmentation} of {Independently}
  {Moving} {Objects} in {Video},'' \emph{Doctoral Dissertations}, Mar. 2020.

\bibitem{zingg_mav_2010}
S.~Zingg, D.~Scaramuzza, S.~Weiss, and R.~Siegwart, ``{MAV} navigation through
  indoor corridors using optical flow,'' in \emph{2010 {IEEE} {Intl.} {Conf.}
  on {Rob.} and {Autom.}}, May 2010, pp. 3361--3368, iSSN: 1050-4729.

\bibitem{hartley_multiple_2004}
R.~Hartley and A.~Zisserman, \emph{Multiple {View} {Geometry} in {Computer}
  {Vision}}, 2nd~ed.\hskip 1em plus 0.5em minus 0.4em\relax Cambridge:
  Cambridge University Press, 2004.

\bibitem{antonelli_bayesian_2014}
M.~Antonelli, A.~P. del Pobil, and M.~Rucci, ``\BIBforeignlanguage{en}{Bayesian
  multimodal integration in a robot replicating human head and eye
  movements},'' in \emph{\BIBforeignlanguage{en}{2014 {IEEE} {Intl.} {Conf.} on
  {Rob.} and {Autom.} ({ICRA})}}.\hskip 1em plus 0.5em minus 0.4em\relax Hong
  Kong, China: IEEE, May 2014, pp. 2868--2873.

\bibitem{duran_robot_2020}
A.~J. Duran and A.~P. del Pobil, ``\BIBforeignlanguage{en}{Robot depth
  estimation inspired by fixational movements},''
  \emph{\BIBforeignlanguage{en}{IEEE Trans. on Cognitive and Developmental
  Systems}}, pp. 1--1, 2020.

\bibitem{santini_active_2007}
F.~Santini and M.~Rucci, ``\BIBforeignlanguage{en}{Active estimation of
  distance in a robotic system that replicates human eye movement},''
  \emph{\BIBforeignlanguage{en}{Robotics and Autonomous Systems}}, vol.~55,
  no.~2, pp. 107--121, Feb. 2007.

\bibitem{raviv_quantitative_1991}
D.~Raviv, ``A quantitative approach to camera fixation,'' in \emph{1991 {IEEE}
  {Computer} {Society} {Conference} on {Computer} {Vision} and {Pattern}
  {Recognition} {Proceedings}}, June 1991, pp. 386--392, iSSN: 1063-6919.

\bibitem{jain_peripheral_1996}
I.~Thomas, E.~Simoncelli, and R.~Bajcsy, ``\BIBforeignlanguage{en}{Peripheral
  {Visual} {Field}, {Fixation} and {Direction} of {Heading}},'' in
  \emph{\BIBforeignlanguage{en}{Exploratory {Vision}}}, R.~C. Jain, M.~S.
  Landy, L.~T. Maloney, and M.~Pavel, Eds.\hskip 1em plus 0.5em minus
  0.4em\relax New York, NY: Springer New York, 1996, pp. 169--189, series
  Title: Springer Series in Perception Engineering.

\bibitem{chaumette_visual_2006}
F.~Chaumette and S.~Hutchinson, ``\BIBforeignlanguage{English}{Visual servo
  control, part {I}: {Basic} approaches},''
  \emph{\BIBforeignlanguage{English}{IEEE Robotics \& Automation Magazine}},
  vol.~13, no.~4, pp. 82--90, Dec. 2006.

\bibitem{simoncelli_distributed_1993}
E.~P. Simoncelli, ``\BIBforeignlanguage{eng}{Distributed representation and
  analysis of visual motion},'' Thesis, M.I.T., 1993, accepted:
  2005-08-15T21:18:23Z.

\bibitem{simoncelli_design_1994}
E.~Simoncelli, ``Design of multi-dimensional derivative filters,'' in
  \emph{Proceedings of 1st {Intl.} {Conf.} on {Image} {Processing}}, vol.~1,
  Nov. 1994, pp. 790--794 vol.1.

\end{thebibliography}

%

\end{document}